# ACTION CHUNKING WITH TRANSFORMERS FOR IMAGE-BASED SPACECRAFT GUIDANCE AND CONTROL

Alejandro Posadas-Nava; Andrea Scorsoglio; Luca Ghilardi; Roberto Furfaro; Richard Linares

We present an imitation learning approach for spacecraft guidance, navigation, and control (GNC) that achieves high performance from limited data. Using only 100 expert demonstrations, equivalent to 6,300 environment interactions, our method, which implements Action Chunking with Transformers (ACT), learns a control policy that maps visual and state observations to thrust and torque commands. ACT generates smoother, more consistent trajectories than a meta-reinforcement learning (meta-RL) baseline trained with 40 million interactions. We evaluate ACT on a rendezvous task: in-orbit docking with the International Space Station (ISS). We show that our approach achieves greater accuracy, smoother control, and greater sample efficiency.

## INTRODUCTION

The development of autonomous spacecraft guidance, navigation, and control (GNC) systems represents one of the most significant challenges in modern space exploration. The nature of space missions naturally necessitates on-board autonomy for two fundamental reasons: (1) mission objectives place constraints on how a spacecraft operates. Missions with goals that are unsuited to traditional ground-controlled operations require autonomy to ensure success. (2) System and environmental constraints, which occur when a spacecraft experiences novel or unpredictable environments, or when communication delays prohibit adequate responses from the spacecraft.[1]

Conventional GNC systems contain three principal components: position determination, attitude determination, and a control system. Each component is designed in context of a spacecraft's mission set. For example, Earth-orbiting satellites use the global positioning system (GPS) for position determination, whereas a deep space probe depends on the deep space network (DSN) and an onboard radio transponder to determine its position.[2] Similarly, highly maneuverable spacecraft, which require precise attitude determination, carry a suite of sensors such as star trackers, magnetometers, and gyros.[2] Both position and attitude determination enable the spacecraft to make control decisions. Spacecraft controllers are based on control laws, predicated on an explicit dynamics model, and designed for specific operational needs.[3] Such controllers are susceptible to performance degradation due to model inaccuracies and rely on precise state estimation.[4] Autonomous systems provide an alternative to traditional controllers by demonstrating improved tolerance to inaccuracies in state estimation.[5] Recent advances in deep reinforcement learning (DRL), meta-reinforcement learning (meta-RL), and imitation learning have shown promise in image-based spacecraft guidance tasks, such as powered descent, allowing policies to learn directly from sensory inputs without relying on explicit dynamics models.[6,7] Meta-RL and RL methods focus on enabling agents to rapidly adapt to new tasks by leveraging previous learning experiences, but require large amounts of data[8] and suffer control "chatter".[9] Previous works, to the best of our knowledge, have used imitation learning for spacecraft control, but never in an end-to-end learning paradigm or with the use of visual navigation.[7,10,11]

---

*Researcher, AeroAstro, MIT, 77 Massachusetts Avenue, Cambridge, MA 02139-4307.
†Postdoctoral Research Associate, SSEL, 1200 E University Blvd, Tucson, AZ 85721
‡Graduate Research Assistant, SSEL, 1200 E University Blvd, Tucson, AZ 85721
§Professor, SSEL, 1200 E University Blvd, Tucson, AZ 85721
¶Associate Professor and Rockwell International Career Development Professor, AeroAstro, MIT, 77 Massachusetts Avenue, Cambridge, MA 02139-4307.

This paper presents a hybrid framework that leverages demonstrations generated by a meta-RL system to train an imitation learning (IL) algorithm. IL enables an agent to acquire task-specific behavior directly from expert demonstrations.[12,13] In our approach, we apply Action Chunking with Transformers (ACT), an IL algorithm that learns spacecraft guidance policies from trajectory data produced by a meta-RL model validated on an optical test bench. A key benefit of ACT is the steep learning curve: With only about $6\,\text{k}$ environment interactions, it outperforms a meta-RL policy, improving distance error on an orbital docking task by $28\%$, trained with approximately $610\,\text{k}$ environment interactions. In addition, ACT mitigates the problem of compounding error in behavioral cloning[14,15] by predicting temporally extended action sequences, called chunks, rather than individual actions. This reduces the effective task horizon and improves temporal consistency,[16] leading to control precision.

In this framework, IL benefits from the strengths of meta-RL and vice versa. If a meta-RL agent generates trajectories with varying performance, only the most successful would be selected as demonstrations for IL training, improving the likelihood of successful policy learning. In contrast, an existing IL policy could serve as a warm start for meta-RL training,[17] illustrating the complementary nature of the two paradigms. IL provides a lightweight and efficient mapping from observations to actions, implicitly capturing system dynamics, safety considerations, and domain-specific constraints.

Our primary contributions are as follows.

1. A hybrid learning framework that uses a meta-reinforcement learning (meta-RL) agent to generate expert trajectories, which are then used to train an Action Chunking Transformer (ACT) policy via imitation learning.

2. A comprehensive evaluation of ACT on a 6-degree-of-freedom (6-DOF) monocular-vision rendezvous and proximity operation (RPO) task: In-Orbit Docking.

**PROBLEM STATEMENT**

**Dynamic Environment**

We begin by formulating the GNC task as a partially observable Markov decision process (POMDP) within the meta-RL framework. This framework provides a principled approach for handling partial observability and learning generalizable policies across a distribution of tasks. The POMDP is defined as in Federici et al.[6] For more details, see Appendix B. At each time step $t$, the spacecraft receives an observation $\mathbf{o}_t$ consisting of images $\mathbf{i}_t$ and sensor data $\mathbf{n}_t$, and executes a 6-DOF action $\mathbf{a}_t = (\mathbf{T}_t, \mathbf{L}_t)$, applying thrust and torque to transition to a new state and obtain reward $r_t$.

Although this meta-RL formulation guides the development of control policies under partial observability, its primary purpose in this work is to serve as a scaffold for generating expert demonstrations. These expert trajectories, collected using policies trained via meta-RL, form the basis for training our ACT model. Thus, while the reinforcement learning phase is essential for acquiring expert demonstrations, our core contribution lies in the development and evaluation of IL techniques that operate purely from such demonstrations.

**Visual Environment**

We simulate high-fidelity visual environments using Blender integrated with OpenAI Gym, producing realistic camera observations synchronized with spacecraft states. The spacecraft is modeled with a body-mounted navigation camera that captures images as it navigates. A Sun-like light source simulates natural illumination by casting parallel rays. Each time step in the simulation updates the positions and orientations of all entities using OpenAI Gym for environment dynamics. These updates are synchronized in real time with the Blender-rendered scene. The navigation camera captures an observation image of size $h \times w$, which reflects a physically accurate view of the environment at that moment. An example of the visualization environment for the docking scenario can be seen in Figure 1.

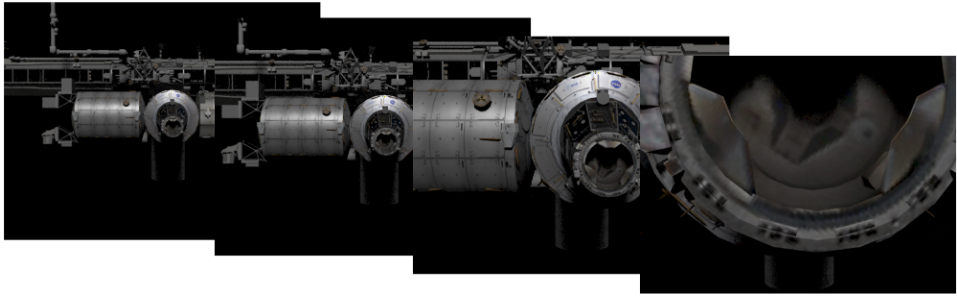

**Figure 1**: Blender visual simulation docking scenario used to generate image-based observations

## METHODOLOGY

We choose Action Chunking with Transformers (ACT) as our visual-state learning model because it is sample-efficient, mitigates the effect of compounding errors, and is computationally lightweight.[16]

**Action Chunking with Transformers**

ACT uses a transformer encoder-decoder architecture combined with a Conditional Variational Autoencoder (CVAE) to model the variability in demonstrations and produce temporally coherent action sequences. The model operates on observations that include RGB images and structured state vectors such as position, velocity, attitude, and angular velocity. The transformer encoder embeds the input observations, specifically the spacecraft state and camera images, into a latent feature space. For each input time step $t$, the observation $o_t$ consists of $o_t = \{\mathbf{I_t}, \mathbf{s_t}\}$, where $\mathbf{I_t}$ is a stack of camera images and $\mathbf{s_t} \in \mathbb{R}^{d_s}$ is a vector representing numerical state data. During training, ACT predicts a sequence of actions $\hat{a}_{t:t+k} \in \mathbb{R}^{k \times d_a}$, where $d_a$ is the dimensionality of the action, and $k$ is the chunk size. The model is trained to minimize the reconstruction loss in these predicted action sequences.

$$\mathcal{L}_{\text{BC}} = \mathbb{E}_{(o_t, a_{t:t+k}) \sim \mathcal{D}} \left[ \| \hat{a}_{t:t+k} - a_{t:t+k} \|_1 \right] \qquad (1)$$

Here, $\mathcal{D}$ is the dataset of expert demonstrations, $a_{t:t+k}$ is the ground-truth action sequence over the chunk window, and $\hat{a}_{t:t+k}$ is the model's prediction. The policy can be expressed as a function.

$$\pi_\theta(a_{t:t+k} \mid o_t, z) = \text{Decoder}\left(\text{Encoder}\left(f(o_t), z\right)\right) \qquad (2)$$

Here, $z \in \mathbb{R}^{d_z}$ is a latent variable sampled from a Gaussian distribution $z \sim \mathcal{N}(\mu, \sigma^2)$ during training, where the mean and variance are produced by a separate CVAE encoder conditioned on both the observation $o_t$ and the target chunk $a_{t:t+k}$.[18] At test time, $z$ is set to the zero vector (*i.e.*, the mean of the prior) to produce deterministic behavior. This latent variable $z$ captures stylistic or multi-modal variations in expert demonstrations. At inference time, the policy outputs a new chunk of $k$ actions at every time step. However, rather than executing only the first action of each chunk, ACT applies *temporal ensembling* to produce smooth trajectories. This method blends overlapping predicted chunks using an exponentially weighted average over the most recent $k$ predictions.

$$a_t = \frac{1}{Z_t} \sum_{i=0}^{k-1} w_i \hat{a}_t^{(i)}, \quad \text{where} \quad w_i = e^{-mi}, \quad Z_t = \sum_{i=0}^{k-1} w_i \tag{3}$$

Here, $\hat{a}_t^{(i)}$ is the predicted action for time $t$ from the $i$-th most recent chunk, $m$ is the temporal ensemble decay coefficient, and $Z_t$ normalizes the weights so that they sum to 1. This ensembling mitigates abrupt changes between consecutive predictions and enhances temporal consistency, which is particularly important in spacecraft GNC scenarios where noisy or discontinuous control signals can degrade stability or lead to unsafe behaviors. The use of a CVAE allows ACT to learn a distribution over valid action sequences conditioned on the current state, improving its ability to model the multi-modal nature of demonstrations.

**Implementation Details**

We implement the ACT model using a transformer encoder-decoder architecture designed to process high-dimensional image observations from multiple camera views alongside numerical spacecraft state data. Expert trajectory data used to train ACT policies are generated by a meta-RL agent deployed in the same environment.[6] These trajectories include spacecraft state information: position, velocity, attitude, and angular velocity, with corresponding actions: thrust and torque.

During training, the encoder, seen in Figure 2a, receives a combination of trajectory samples including spacecraft state, actions, and a learned "[CLS]" token, which serves as a summary embedding for encoding action chunks. The trained "[CLS]" token outputs a latent style variable $z$, representing the underlying structure of the expert behavior. To fuse image and state inputs, the model processes visual observations from each camera through a ResNet-based convolutional backbone to extract spatial feature maps. These maps are projected to a fixed hidden dimension using a $1 \times 1$ convolution and augmented with 2D sinusoidal positional encodings to preserve spatial structure. The resulting image embeddings are flattened and concatenated across camera views to form a sequence of visual tokens. Meanwhile, the spacecraft state is embedded with a linear projection and enriched with learned 1D positional encodings to yield a compact state token. The image tokens, state token, and latent style vector are stacked into a unified sequence and passed through a multi-layer transformer encoder, which models spatial-temporal correlations via self-attention.

The fused output of the encoder is then passed to a transformer decoder, which predicts action chunks. The decoder receives a set of learned positional queries, one per action timestep, and first applies self-attention to model temporal dependencies between predicted actions. Then it uses cross-attention to condition each action prediction on the encoder's fused representation of the current observation, enabling the model to attend selectively to spatial and contextual features relevant to the task. This architecture ensures that generated actions are both temporally consistent and grounded in visual and dynamic context, facilitating precise control. The decoder outputs action chunks conditioned on the observation and latent style variable $z$, as shown in Figure 2b.

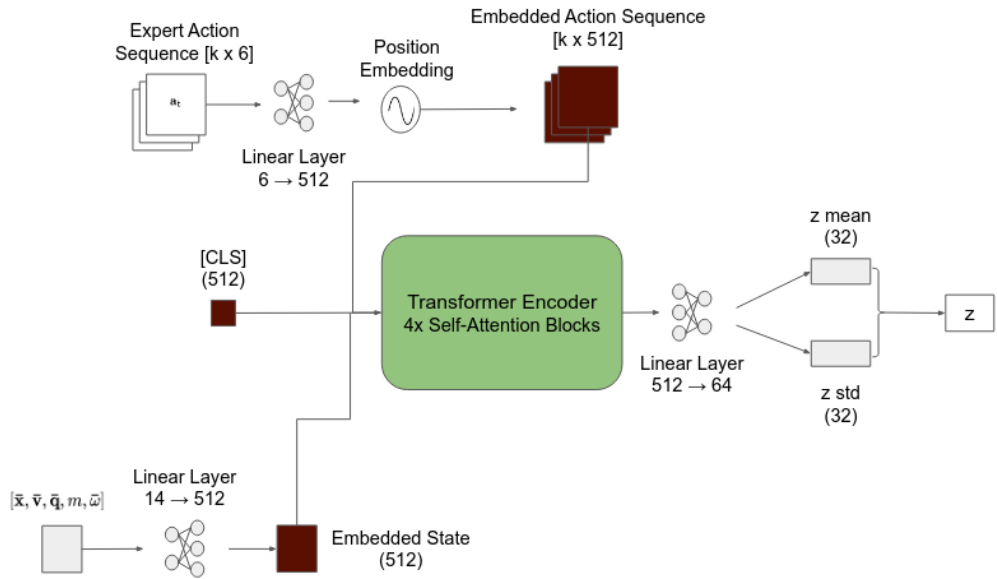

(a) Inferring the style variable $z$

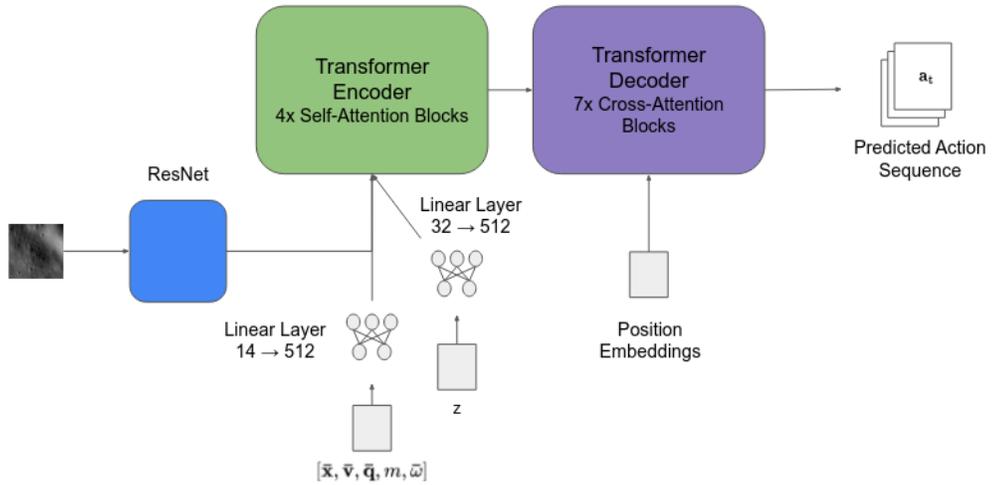

(b) Predicting an action sequence.

**Figure 2**: ACT model architecture.

## EVALUATION AND RESULTS

We tested ACT in an orbital docking scenario. We compared the performance of the ACT policy with the performance of the RL network in simulation. To enable a one-to-one performance comparison with prior work, we replicated the in-orbit docking environment introduced by Federici et al.[6] The environment, observation stack (image + state), control limits, and evaluation protocol remain unchanged. Our method achieved substantially higher data efficiency: While the meta-RL baseline required $63,000$ episodes ($\approx 40$ million environment interactions), ACT learned an effective policy from only 100 demonstrations ($\approx 6300$ environment interactions), as summarized in Table 1.

Table 1: Average sample efficiency across models

| Model | Number of episodes | Number of environment interactions |
|---|---|---|
| RL | $63,000$ | $40,000,000$ |
| ACT | $100$ | $6,350$ |

### In-Orbit Docking

We reused the image–based docking environment of Federici et al.[6] to allow a direct comparison. A chaser equipped with a body-mounted navigation camera must rendezvous with an ISS model rendered in Blender. Relative translation follows the Hill–Clohessy–Wiltshire equations in a Local-Vertical-Local-Horizon frame centered on the docking port, with the station on a $409\,\mathrm{km}$ circular orbit. Rotational motion is integrated with Euler equations, resulting in a six 6-DOF state. At a variable time step ($\Delta t \approx 0.89 \pm 0.13$ s), the agent receives a $480 \times 640$ RGB image plus the current position, velocity, attitude, and angular-rate vector. The policy outputs cumulative thrust and torque commands.

The chaser begins $25\,\mathrm{m}$ away from the ISS along the negative V-bar, and an initial attitude such that the camera points at the docking port. We ran two evaluation procedures, which created evaluation datasets each with 100 episodes. In the first trial, the chaser's initial conditions were set to match the initial conditions seen in training. In the second test, we tested the ability of the policy to generalize by randomizing the initial conditions within a larger dispersion area. Both sets of initial conditions are shown in Table 2.

Table 2: Initial position bounds across episodes

| Axis | Same (m) | Random (m) |
|---|---|---|
| X | $-1$ to $1$ | $-2.5$ to $2.5$ |
| Y | $-26$ to $-24$ | $-27.5$ to $-22.5$ |
| Z | $-1$ to $1$ | $-2.5$ to $2.5$ |

When using the initial conditions seen during training, the ACT controller outperformed RL, achieving greater precision and accuracy for both distance and velocity. In particular, ACT came within $80\,\mathrm{cm}$ of the docking port on $43\%$ of the runs, while RL never came closer than $87\,\mathrm{cm}$. ACT got within $18\,\mathrm{cm}$ of the port on its best run, almost five times better than the RL network. We compare the performance of the ACT and RL policies in Table 3. Two runs, falling $2.50\,\mathrm{m}$ and $2.24\,\mathrm{m}$ away from the docking port, skewed the performance of the ACT controller. However, ACT outperforms RL in general, achieving a lower average distance and speed at the end of each episode.

The trajectories of both evaluation procedures are shown in Figure 3. Figure 3a illustrates the trajectories of the chaser from starting states equal to the RL data used in training. Figure 3b illustrates the trajectories of the chaser from random starting positions dispersed within a 4 m cube. In this case, the controller came within 4 m of the port with a 37% success rate, demonstrating some generalization, but only came within 2 m of the docking port on 5% of the runs.

Table 3: RL vs. ACT performance on the docking scenario

| Model | $\|r_K\|$ (m) | | | | $\|v_K\|$ (m/s) | | | |
|---|---|---|---|---|---|---|---|---|
| | Mean | 75% | 95% | 99% | Mean | 75% | 95% | 99% |
| RL | 1.391 | 1.557 | 1.738 | 1.810 | 0.092 | 0.100 | 0.110 | 0.1241 |
| ACT | 0.992 | 1.326 | 1.763 | 2.248 | 0.086 | 0.112 | 0.141 | 0.159 |

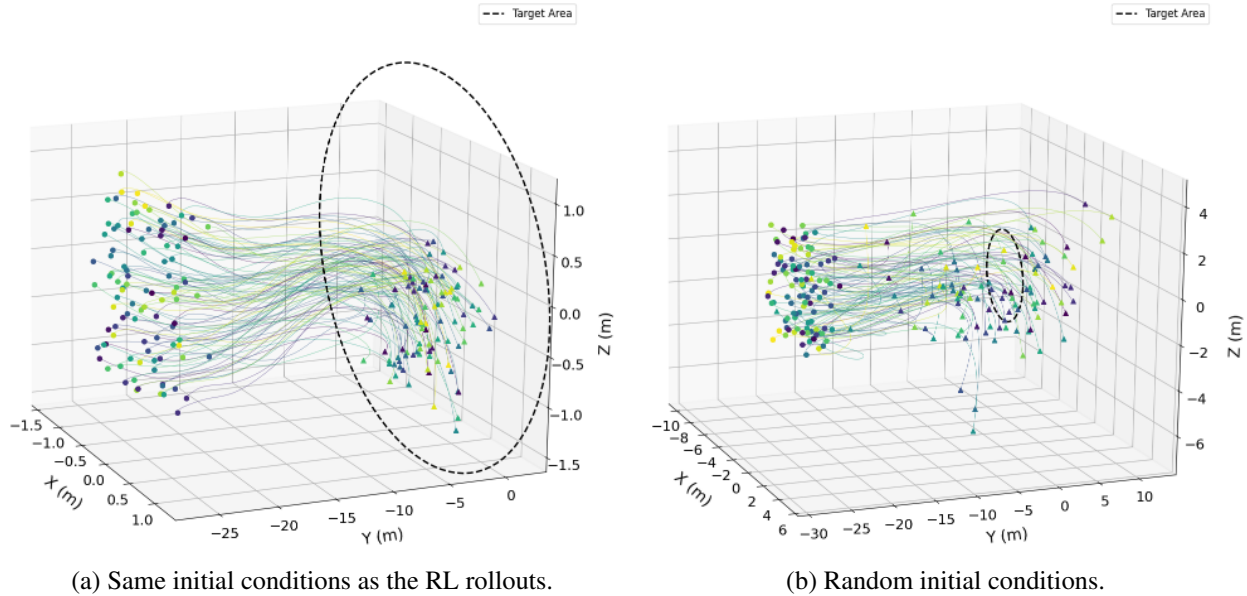

(a) Same initial conditions as the RL rollouts.  (b) Random initial conditions.

Figure 3: Comparison of 100 trajectories produced by the ACT controller under (a) identical initial conditions and (b) randomized initial conditions. Circles mark start positions, triangles mark end positions, and the ring shows a 2 m docking radius.

Next, we compared the control smoothness between both controllers. We computed the mean smoothness score per episode using the $L_2$ norm of successive action differences for both the ACT and RL datasets.

$$\Delta a_t = \|a_{t+1} - a_t\|_2 \qquad (4)$$

Here, $a$ denotes the combined thrust and torque action vector issued by the controller at time step $t$. The resulting scalar values were averaged across each episode, yielding one smoothness score per episode. The ACT controller produced significantly smoother action trajectories compared to the RL baseline. A Welch's two-sample $t$-test was used to compare the group means. The result, shown in Table 4, was statistically significant, with $t(135) = -94.77$ and $p < 0.001$ (the corresponding two-tailed probability under the null hypothesis of equal means), which provides strong evidence that ACT generates smoother control actions than RL in this setting.

Both smoothness distributions are shown in Figure 4. The much tighter and lower mean distribution for ACT indicates significantly smoother control commands compared to the RL controller.

Table 4: Smoothness statistics for ACT vs. RL controllers (lower is smoother)

| Model | Mean ($M$) | Std. Dev. ($SD$) | Sample Size ($N$) | $t$-statistic | $p$-value |
|---|---|---|---|---|---|
| RL | 9.39 | 0.74 | 100 | $t(135) = -94.77$ | $< 0.001$ |
| ACT | 1.71 | 0.32 | 100 | | |

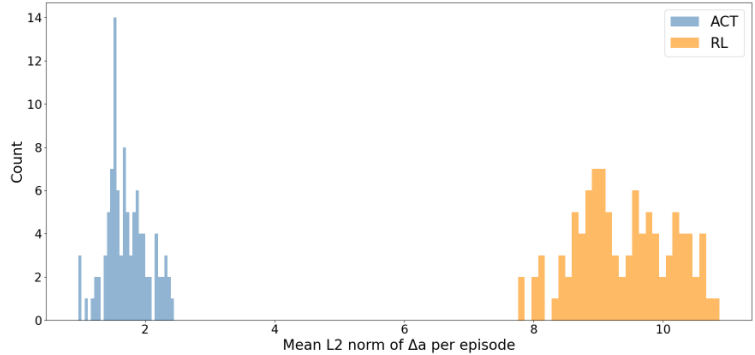

**Figure 4**: Distribution of action-step magnitudes ($\|\Delta a_t\|_2$) under the baseline RL and ACT rollouts.

The smoothing effect was the result of action chunking, which reduces the horizon of the trajectory, and temporal ensembling. Figure 5, clearly shows how jagged RL actions are transformed into smoother and more continuous control outputs.

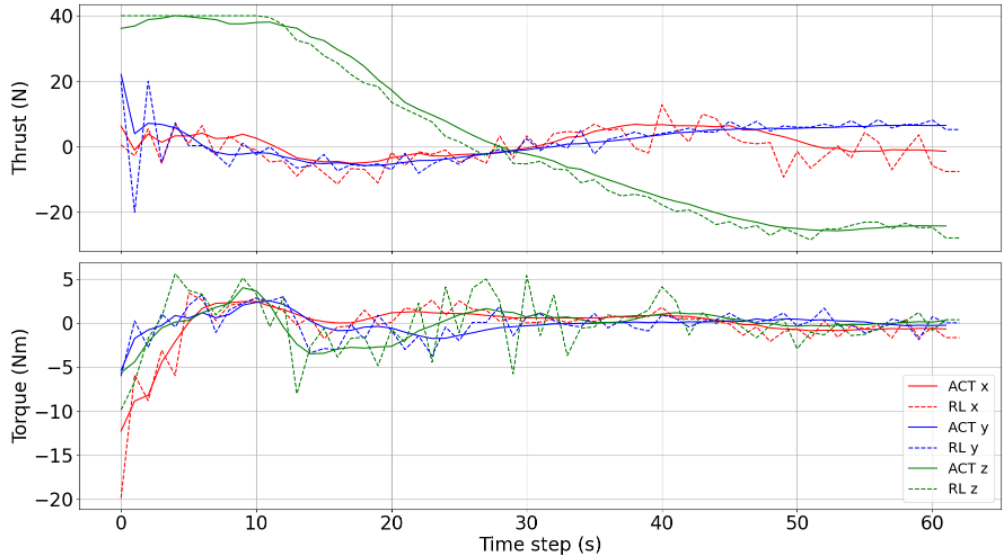

**Figure 5**: Comparison of actions between ACT and RL policies from a sampled trajectory.

Furthermore, the ACT controller captures trajectory modes that lead to consistent and repeatable behavior. This idea is reinforced in Figure 6, which shows contour plots rendered as topographic maps: darker regions denote high visitation density across all episodes. ACT produced several narrow, high peaks, showing a few repeatable approach modes, whereas the RL policy yields a single broad, low amplitude hill, indicating more variability in its trajectories. These high-density corridors align with the smoother control profiles observed when using ACT and suggest a consistent convergence to stable maneuver strategies.

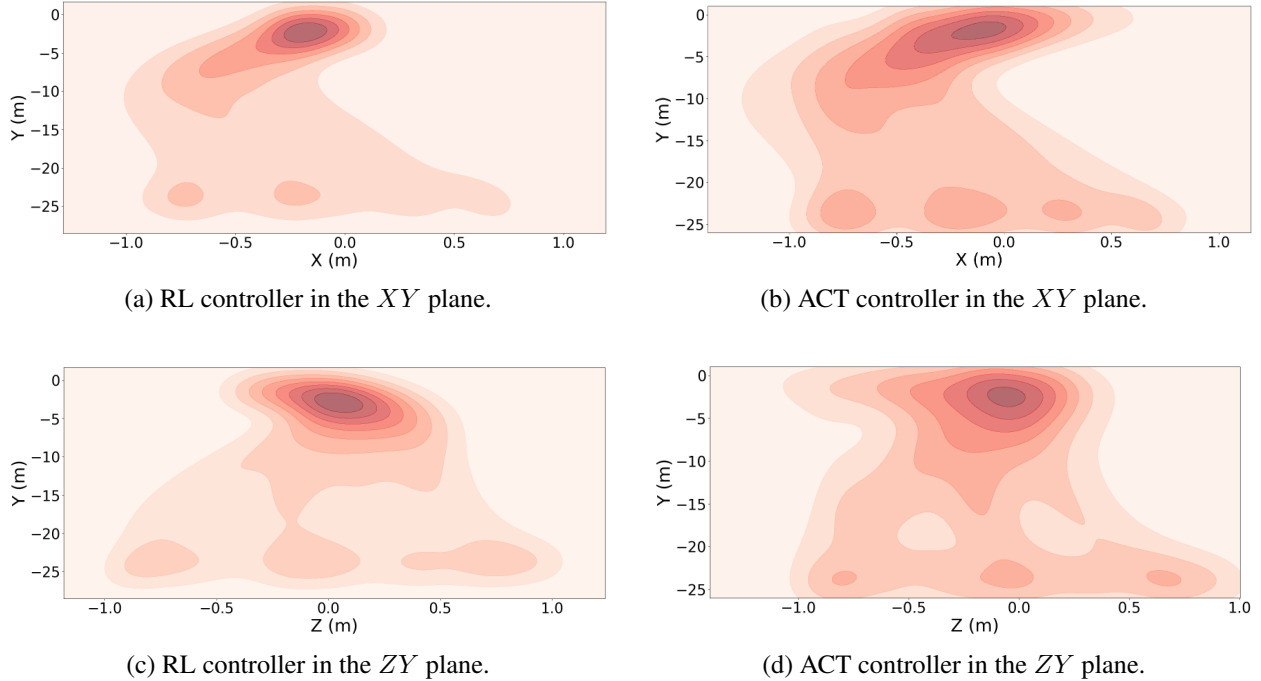

**Figure 6**: Trajectory-visitation density heatmaps for the RL (left column) and ACT (right column) controllers in the $XY$ and $ZY$ planes. Darker regions denote higher visitation frequency.

These results confirm that ACT not only achieves better docking precision, but also generates structured, repeatable control behavior with only 100 sample trajectories.

### LIMITATIONS

Although our approach improves performance, several limitations remain. First, the ACT controller, in its best-case scenario, does not consistently meet the $10\,\text{cm}$ accuracy threshold required for ISS docking operations.[19] This limitation is attributed in part to the quality of the demonstration data. Second, we did not perform a systematic analysis of how various training hyperparameters, such as chunk size, action step resolution, and number of training iterations, impact policy performance. Lastly, our current approach trains ACT using entire trajectories, which may not be optimal. A promising direction for future work is to focus training on lower-variance segments of the trajectory where finer control is required. This could potentially mitigate the errors associated with long-horizon predictions and randomized initializations.

### CONCLUSION

This work presents a hybrid learning pipeline that uses meta-RL to generate expert trajectories, which are then distilled into deployable control policies using ACT. By predicting sequences of control actions and incorporating temporal ensembling, ACT produces smooth, precise maneuvers suitable for spacecraft guidance scenarios. The model is trained using, on average, only $6,530$ environment interactions compared to $40$ environment interactions required to train the benchmark meta-RL network. Our approach achieves a $29\%$ increase in mean precision for the in-orbit docking task and produces, on average, $82\%$ smoother control outputs. Our results demonstrate that ACT, trained on expert demonstrations from a validated meta-RL agent, can accurately map image and state observations to thrust and torque commands with greater sample efficiency, accuracy, and smoothness than meta-RL alone.

# APPENDIX A

## Smoothness metric and statistical analysis

To quantify how smoothly each controller changes its commanded actions, we computed the Euclidean difference using Equation 4 for both thrust and torque at each time step and then averaged this value over the entire episode. This produces a smoothness score per episode. With 100 independent episodes per policy, we obtained two samples of size $N_{\text{IL}} = N_{\text{RL}} = 100$.

**Assumption checks.** Before comparing group means we verified the prerequisites for parametric testing:

1. *Independence* is guaranteed by the experimental design (each episode is an independent simulation with an identical initial state but separate random seeds).

2. *Normality* of the episode-level scores was assessed with Shapiro–Wilk tests and Q–Q plots; in both groups the null hypothesis of normality was not rejected ($p_{\text{ACT}} = 0.094$, $p_{\text{RL}} = 0.085$).

3. *Homogeneity of variances* was examined with Levene's test, which indicated heteroscedasticity ($F = 70.30$, $p < 10^{-14}$).

**Choice of test.** Because the assumption of normality held while the variances differed significantly, we used Welch's two–sample $t$-test (unequal-variance $t$-test) to compare the mean smoothness scores. Welch's test is designed to be used when the variances of two groups are unequal and is therefore recommended as default when Levene's test is significant. All tests were two-tailed with $\alpha = 0.05$.

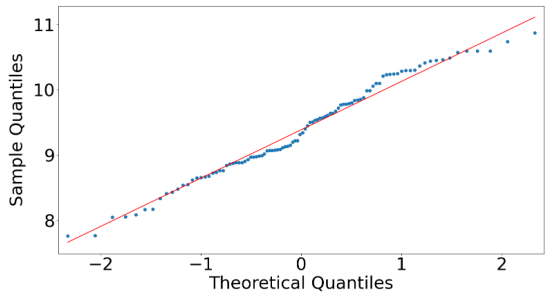
(a) Q–Q plot for the RL policy (rollout-level $\Delta a$ $L_2$ norms).

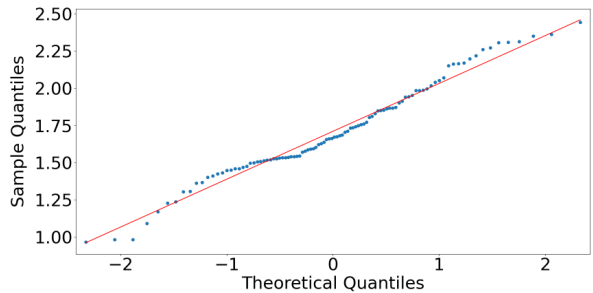
(b) Q–Q plot for the ACT policy (rollout-level $\Delta a$ $L_2$ norms).

**Figure 7**: Q–Q plots assessing the normality of rollout-level smoothness metrics ($L_2$ norm of successive action differences) for the RL and ACT controllers. Points closely following the 45° reference line indicate that both distributions deviate only mildly from normality, supporting the use of parametric tests.

## APPENDIX B

Paritally observable Markov Decision Process tuple: $\langle \mathcal{S}, \mathcal{A}, \mathcal{O}, T, R, \Omega, \gamma \rangle$, where $\mathcal{S}$ is the hidden state space, $\mathcal{A}$ the action space, $\mathcal{O}$ the observation space, $T$ the transition dynamics, $R$ the reward function, $\Omega$ the observation model, and $\gamma$ the discount factor.